\documentclass{article}

\usepackage{subfigure}
\usepackage{listings}
\usepackage{extarrows}
\usepackage{courier}

\usepackage{color}
\usepackage[table,xcdraw]{xcolor}
\usepackage{epsfig}
\usepackage{graphicx}

\usepackage{adjustbox}
\usepackage{array}
\usepackage{booktabs}
\usepackage{colortbl}
\usepackage{float,wrapfig}
\usepackage{hhline}
\usepackage{multirow}
\usepackage{amsmath,amsfonts,amssymb}
\usepackage{bm}
\usepackage{nicefrac}
\usepackage{microtype}

\usepackage{changepage}
\usepackage{extramarks}
\usepackage{fancyhdr}
\usepackage{lastpage}
\usepackage{setspace}
\usepackage{soul}
\usepackage{xspace}

\usepackage[pagebackref=true,breaklinks=true,colorlinks,citecolor=gray]{hyperref}
\usepackage{url}

\usepackage{algorithm, algorithmic}
\usepackage[ruled,vlined,noresetcount,algo2e]{algorithm2e}
\usepackage{enumerate}
\usepackage{todonotes} %
\usepackage{pdfpages}
\newcolumntype{L}[1]{>{\raggedright\let\newline\\\arraybackslash\hspace{0pt}}m{#1}}
\newcolumntype{C}[1]{>{\centering\let\newline\\\arraybackslash\hspace{0pt}}m{#1}}
\newcolumntype{R}[1]{>{\raggedleft\let\newline\\\arraybackslash\hspace{0pt}}m{#1}}

\newcommand{\sect}[1]{Section~\ref{#1}}

\newcommand{\fig}[1]{Figure~\ref{#1}}

\newcommand{\tbl}[1]{Table~\ref{#1}}

\newcommand{\ignorethis}[1]{}

\makeatletter
\DeclareRobustCommand\onedot{\futurelet\@let@token\@onedot}
\def\@onedot{\ifx\@let@token.\else.\null\fi\xspace}

\def\iid{i.i.d\onedot}
\def\eg{e.g\onedot} 
\def\ie{i.e\onedot} 
 
\def\etc{etc\onedot} 
\def\wrt{w.r.t\onedot} 
 \def\aka{a.k.a\onedot}
\makeatother

\definecolor{MyDarkBlue}{rgb}{0,0.08,1}
\definecolor{MyDarkGreen}{rgb}{0.02,0.6,0.02}
\definecolor{MyDarkRed}{rgb}{0.8,0.02,0.02}
\definecolor{MyDarkOrange}{rgb}{0.40,0.2,0.02}
\definecolor{MyPurple}{RGB}{111,0,255}
\definecolor{MyRed}{rgb}{1.0,0.0,0.0}
\definecolor{MyGold}{rgb}{0.75,0.6,0.12}
\definecolor{MyDarkgray}{rgb}{0.66, 0.66, 0.66}

\newcommand{\modelfull}{Neurally-Guided Structure Inference\xspace}
\newcommand{\model}{NG-SI\xspace}
\newcommand{\myparagraph}[1]{\vspace{-5pt}\paragraph{#1}}

\usepackage{amsmath,amsfonts,bm}

\def\eqref#1{equation~\ref{#1}}

\def\1{\bm{1}}

\DeclareMathAlphabet{\mathsfit}{\encodingdefault}{\sfdefault}{m}{sl}
\SetMathAlphabet{\mathsfit}{bold}{\encodingdefault}{\sfdefault}{bx}{n}

\usepackage[accepted]{icml2019}

\icmltitlerunning{Neurally-Guided Structure Inference}

\begin{document}
\twocolumn[
\icmltitle{Neurally-Guided Structure Inference}

\icmlsetsymbol{equal}{*}

\begin{icmlauthorlist}
\icmlauthor{Sidi Lu}{equal,sjtu}
\icmlauthor{Jiayuan Mao}{equal,mitcsail,tsinghua}
\icmlauthor{Joshua B. Tenenbaum}{mitcsail,mitbcs,mitcbmm}
\icmlauthor{Jiajun Wu}{mitcsail}
\end{icmlauthorlist}

\icmlaffiliation{sjtu}{Shanghai Jiao Tong University,}
\icmlaffiliation{mitcsail}{MIT CSAIL,}
\icmlaffiliation{tsinghua}{IIIS, Tsinghua University,}
\icmlaffiliation{mitbcs}{Department of Brain and Cognitive Sciences, MIT,}
\icmlaffiliation{mitcbmm}{Center for Brains, Minds and Machines (CBMM), MIT}

\icmlcorrespondingauthor{Sidi Lu}{steve\_lu@apex.sjtu.edu.cn}
\icmlcorrespondingauthor{Jiajun Wu}{jiajunwu@mit.edu}

\icmlkeywords{Structure Inference, Guided Search}

\vskip 0.3in
]

\printAffiliationsAndNotice{\icmlEqualContribution} %
{\par
Project Page: \url{http://ngsi.csail.mit.edu}
}

\begin{abstract}
    Most structure inference methods either rely on exhaustive search or are purely data-driven. Exhaustive search robustly infers the structure of arbitrarily complex data, but it is slow. Data-driven methods allow efficient inference, but do not generalize when test data have more complex structures than training data. In this paper, we propose a hybrid inference algorithm, the Neurally-Guided Structure Inference (NG-SI), keeping the advantages of both search-based and data-driven methods. The key idea of NG-SI is to use a neural network to guide the hierarchical, layer-wise search over the compositional space of structures. We evaluate our algorithm on two representative structure inference tasks: probabilistic matrix decomposition and symbolic program parsing. It outperforms data-driven and search-based alternatives on both tasks.
\end{abstract}
\vspace{1em}
\section{Introduction}

\begin{figure*}[t]
    \centering
    \includegraphics[width=\textwidth]{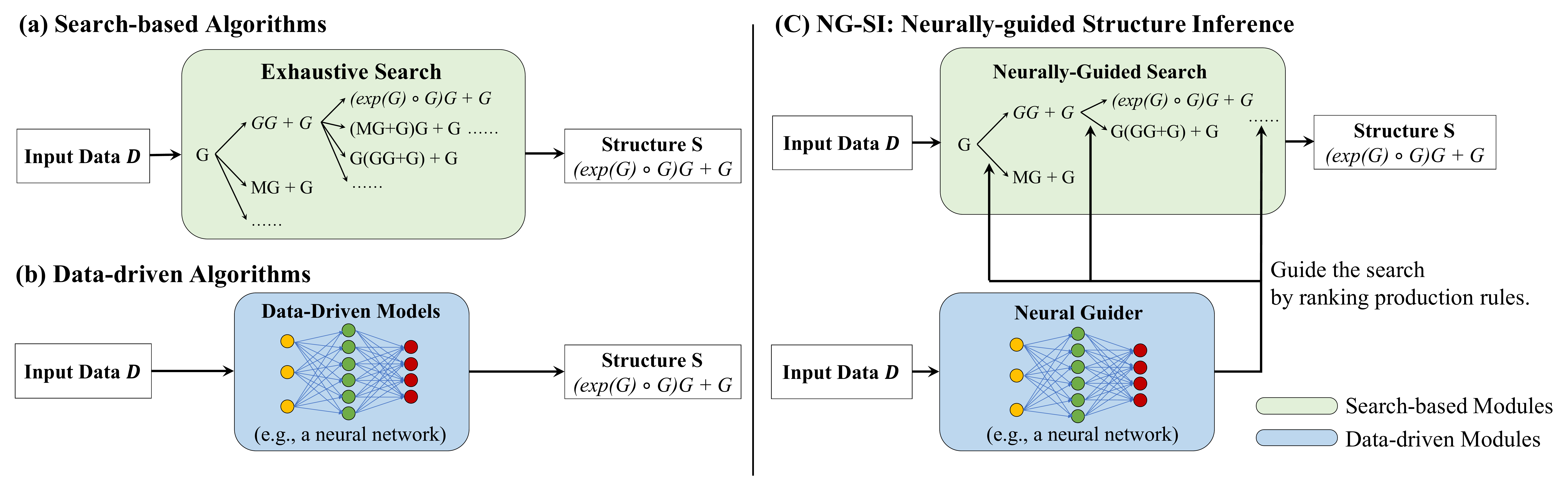}
    \caption{The illustrative flowcharts for (a) search-based algorithms, (b) data-driven algorithms, and (c) the proposed \model. \model uses a neural network (a data-driven module) to guide a hierarchical, layer-wise search process. It outperforms search-based and data-driven alternatives in both inference robustness and efficiency.}
    \label{fig:teaser}
    \vspace{-15pt}
\end{figure*} 
At the heart of human intelligence is the ability to infer the structural representation of data. Looking at hand-written digits from the MNIST dataset \cite{LeCun1998Gradient}, we humans effortlessly group them by digits and extract key features such as angles and thickness of strokes.
Throughout the years, researchers have developed many practically useful compositional structures.
A non-exhaustive list includes low-rank factorization~\citep{mnih2008probabilistic}, clustering, co-clustering~\citep{kemp2006learning}, binary latent factors~\citep{ghahramani2006infinite}, sparse coding~\citep{olshausen1996emergence,berkes2008sparsity}, and dependent GSM~\citep{karklin2009emergence}.

An emerging research topic is to discover appropriate structures from data automatically~\citep{Kemp2008discovery,Grosse2012Exploiting}. These methods are motivated by a key observation: structures can be expressed as hierarchical compositions of primitive components. Given the set of primitives and the production rules for composition (\aka a domain-specific language), they use various inference algorithms to find the appropriate structure.

These inference algorithms can be roughly divided into two categories: search-based and data-driven. Search-based algorithms (\fig{fig:teaser}a) look for the best structure by an exhaustive search over the compositional space of possible structures. Candidate structures are ranked by expert-designed metrics. Such search routines are robust in inferring structures of arbitrary complexity, but they can be prohibitively slow for complex structures.

By contrast, data-driven algorithms (\fig{fig:teaser}b) learn to infer structures based on annotated data. Among them, deep neural networks enable efficient amortized inference: by learning from past inferences, future inferences run faster. Nevertheless, data-driven methods tend to overfit to training examples and fail to generalize to test data with more complex structures.

In this paper, we propose the \modelfull (\model, see \fig{fig:teaser}c), a hybrid inference algorithm that integrates the advantages of both search-based and data-driven approaches. In particular, it uses a neural network learning to guide a hierarchical, layer-wise search process.
For each layer, the neural guider outputs a probability distribution over all possible production rules that can be applied. Based on this ranking, only a small number of rules are evaluated. This remarkably reduces the number of nodes in the search tree.

We evaluate \model on two representative structure inference tasks: probabilistic matrix decomposition and symbolic program parsing. It outperforms data-driven and search-based alternatives in both inference robustness and efficiency.

\section{Structure Inference}

In this section, we formally introduce the task of inferring hierarchical structures from data.

\subsection{Problem Formulation}
We want to jointly infer the structure $S$ and its associated parameters $T$ that best explain the observed data $D$. 
As a motivating example, the collection of hand-written digits $D$ from the MNIST dataset can be clustered by their digits or the thickness of their strokes; they form the parameter space of $T$. The structure $S$ here is ``clustering'': $MG+G$, where the $M$ is a multinomial variable modeling the cluster labels, and the $G$'s are Gaussian variables modeling the cluster centers and the data noise. Here $S$ and $T$ form the structural representation of $D$.

Simple structures can be hierarchically composed into a more complex one. \fig{fig:running-example} shows an illustrative inference of the structure $S=$~(MG + G)G + G from the data matrix $D$ \citep{Grosse2012Exploiting}. The matrix has a low-rank structure in columns, while the first factorized component has a finer structure of row clustering.
We use a domain-specific language (DSL) to represent the compositional space of possible structures $\{S\}$. Throughout this paper, we assume the existence of a context-free grammar for the DSL. Beginning from the start symbol, we can apply arbitrary production rules over non-terminal symbols in any order to generate syntactically correct structures. The terminal symbols in the context-free grammar represent the primitive concepts in the domain, such as cluster labels or centers.

\renewcommand{\myparagraph}[1]{\vspace{-10pt}\paragraph{#1}}

\subsection{Prior Work}
The history of structure inference dates back to \citet{vitanyi1997introduction},
where researchers discussed representing data as (program, input) pairs.
Below, we discuss recent progress on structure inference. Existing methods mostly fall into two categories: search-based and data-driven.

\begin{figure}[t]
\vspace{-10pt}
    \centering
    \begin{align*}
    \begin{bmatrix}
    0 & 1 & 1 & 1 \\
    1 & 1 & 1 & 0 \\
    0 & 1 & 1 & 1 
    \end{bmatrix} \xlongequal{\text{Low-rank}}
    \underset{\text{Structure: GG + G}}{\underline{
    \begin{bmatrix}
    0 & 1 & 1 \\
    1 & 1 & 0 \\
    0 & 1 & 1
    \end{bmatrix} \times \begin{bmatrix}
    1 & 0 & 0 & 0\\
    0 & 1 & 1 & 0 \\
    0 & 0 & 0 & 1
    \end{bmatrix}}}& \\
    \xlongequal{\text{Clustering}}
    \underset{\text{Structure: (MG + G)G + G}}{\underline{
    \left(
    \begin{bmatrix}
    1 & 0 \\
    0 & 1 \\
    1 & 0 \\
    \end{bmatrix} \times
    \begin{bmatrix}
    0 & 1 & 1 \\
    1 & 1 & 0 \\
    \end{bmatrix} \right) \times \begin{bmatrix}
    1 & 0 & 0 & 0\\
    0 & 1 & 1 & 0 \\
    0 & 0 & 0 & 1
    \end{bmatrix}}}&\\
    \end{align*}
    \vspace{-40pt}
    \caption{An example of the hierarchical matrix decomposition.}
    \vspace{-15pt}
    \label{fig:running-example}
\end{figure}

\myparagraph{Search-based structure inference.}
Search-based algorithms (\fig{fig:teaser}a) infer the structure of data with an exhaustive search over the compositional space of possible structures.  The compositional space for structures is defined by domain experts.
\citet{Kemp2008discovery} composed a graph where the edges represent the dependencies between data entries.
\citet{Grosse2012Exploiting} proposed a context-free grammar for probabilistic matrix decomposition.
The grammar is powerful enough to encompass a large collection of Bayesian models for matrices: binary latent factors, dependent GSM, \etc. Off-the-shelf matrix decomposition algorithms are used for data decomposition, while a posterior marginal likelihood is used to rank candidate structures.
It also enables a hierarchical search process for structure inference, which significantly reduces the size of the searching space.
\citet{duvenaud2013structure} defined a grammar for generating kernels for Gaussian process models, while \citet{Lloyd2014Automatic} generated natural language descriptions of time series data.
These frameworks require an enumeration of structures or production rules, followed by an evaluation based on the input to rank candidate structures. Thus, they are too slow in inferring complex structures on large-scale data. In this paper, we tackle this issue by introducing a data-driven module that learns to guide the hierarchical, layer-wise search process.

The search-based inference is usually performed in a recursive way~\citep{Grosse2012Exploiting, Kemp2008discovery}. An illustrative example is shown in \fig{fig:running-example}. First, the algorithm expands the start symbol of the grammar with all available production rules; it then decomposes the input data into multiple components, where each component corresponds to one symbol in each of the production rules (\eg, performs low-rank factorization on the original matrix). These components can be further recursively decomposed (\eg, performs clustering on the first component matrix). An expert-designed metric is applied to rank all candidate structures after each expansion. In practice, only top-ranked candidate structures are further expanded. Such algorithms are robust \wrt the complexity of the structures, but are slow for large-scale data.

\myparagraph{Data-driven structure inference.}
Data-driven algorithms (\fig{fig:teaser}b) infer the structure of data by learning from annotated data.
Deep neural networks enable efficient amortized inference: after learning from past inferences, they can efficiently recognize similar structures for test data. Structure inference on sequential data includes {\it sequence labelling}, assigning categorical labels (\eg, part-of-speech tags) for each item in the sequence~\citep{ma2016end}, {\it parsing}, generating graphical structures of input sequences such as dependency trees~\citep{yih2014semantic,chen2014fast}, and {\it sequence-to-sequence translation}~\citep{Sutskever2014Sequence}, which has been further extended to sequence-to-tree learning~\citep{dong2016language} and tree-to-tree learning~\citep{chen2018tree}. This line of research has been adapted to applications such as program synthesis~\citep{parisotto2016neuro, reed2015neural}.
As an representative example, {\it sequence-to-sequence} model \citep{Sutskever2014Sequence} generates the entire structure symbol by symbol. 
However, the symbolic and compositional nature of structures has brought a remarkable difficulty to data-driven approaches: they tend to overfit to training examples and fail to generalize to test data with more complex structures. For example, in sequence encoding, recurrent networks easily fail when tested on longer sequences.
In general, data-driven approaches are more efficient but less robust than search-based alternatives.

\myparagraph{Guided search.} Data-driven models such as neural networks can be used to improve the efficiency of search-based inference. \citet{benevs2011guided} first implemented this idea in procedural modeling, where the task is to reconstruct visual data such as objects or textures using a set of generative rules. In \citet{benevs2011guided}, external data-driven models are used to improve the stability of the procedural modeling. Similarly, \citet{ritchie2016neurally} proposed to embed neural modules into the stochastic algorithm. The neural modules are trained to maximize the likelihood of the outputs generated by sequential Monte Carlo. \citet{menon2013machine} and \citet{Devlin2017Robustfill} proposed to use data-driven methods for program induction. Many algorithms have been proposed to improve the guided search by introducing type information \citep{osera2015type}, learning from multiple demonstrations \citep{Sun2018Neural}, incorporating hierarchically structured traces \citep{fox2018parametrized}, using execution-guided inference \citep{chen2018executionguided,tian2018learning,zohar2018automatic,wang2018robust}, predicting attributes of programs \citep{Balog2017Deepcoder,ellis2018learning}, predicting sketches \citep{lezama2008program,murali2017neural}, leveraging constraint logic programming \citep{zhang2018neural}, or inducing subroutines from existing programs \citep{ellis2018library}.

Unlike these methods, we exploit compositionality in structure inference: we use neural networks to amortize the per-layer inference and keep the hierarchical search process. Each production rule is associated with an expert-designed algorithm to decompose the data into multiple components, and the same inference algorithm could be recursively applied to each of the components. This helps the generalization of the amortized inference. The idea of explicitly incorporating recursion has also been studied in program synthesis~\cite{cai2017making,chen2018towards}. In particular, \citet{kalyan2018neural} have also proposed to guide a hierarchical search by leveraging witness functions to decompose the full program synthesis problem into multiple sub-problems. While their algorithm requires real-world data for training, our neural guider can purely learn from synthetic data and generalize to more complex, real-world structures.

\renewcommand{\myparagraph}[1]{\vspace{-5pt}\paragraph{#1}} 
\section{Neurally-Guided Structure Inference}
\begin{algorithm}[t]
  \SetKwProg{Fn}{Function}{:}{}
  \SetKwFunction{FInner}{Infer}
  \Fn{\FInner{$D$, $Type$}}{
    $rule \leftarrow \texttt{SelectRule}(D, Type)$\;\\
    \For{{\rm each non-terminal symbol} $s$ {\rm in} $rule$}{
        $C_s \leftarrow \texttt{DecomposeData}(D, rule, s)$\;\\
        Replace $s$ in $rule$ with \FInner{$C_s$, $s$}\;
    }
    \Return{$rule$}
  }
\caption{Neurally-Guided Structure Inference}
\label{alg:recursive-inference}
\end{algorithm}

In this paper, we propose the \modelfull (\model, see \fig{fig:teaser}c), a hybrid inference algorithm that integrates the advantages of both search-based and data-driven approaches. Algorithm~\ref{alg:recursive-inference} shows the pseudocode. The algorithm builds the hierarchical structure by recursively choosing the production rule to expand a non-terminal symbol. Meanwhile, a decomposition algorithm, \texttt{DecomposeData}, decomposes the input data $D$ into several components ($\{C_s\}$), whose structure can be recursively inferred by the \texttt{Infer} function. For simplicity, in Algorithm~\ref{alg:recursive-inference}, only one production rule is applied to a non-terminal symbol (determined by \texttt{SelectRule}). One can also extend this greedy algorithm to the beam search-based inference or other variants.

We implement $\texttt{SelectRule}$ as a neural network, namely, the neural guider. The neural guider enables efficient amortized inference at a layer level: it learns to predict the best production rule for a non-terminal symbol in the structure. By contrast, purely search-based approaches need to evaluate all possible rules to select the best one.

We illustrate this idea by a running example of matrix decomposition based on the DSL introduced by \citet{Grosse2012Exploiting}. Consider the simple matrix shown in \fig{fig:running-example}. Note that the matrix does not have a full column rank. Thus, at the first step, the neural guider should select the ``low-rank factorization'' rule and decompose the input data into two components. Next, note that the first decomposed matrix has a clustering structure: all row vectors can be grouped around two centers. Thus, the neural guider should select the ``clustering'' rule at the second step. Such recursive inference may continue for data with more complex structures.

Recursion lies at the core of \model. First, recursion reduces the infinitely compositional search space of structures to the space of available production rules. It also ensures generalization to arbitrarily complex data. Moreover, recursion is a critical inductive bias for the data-driven neural guider. The amortized inference happens at the layer level instead of at the full structure level. This ensures efficient structure inference while allowing combinatorial generalization.

Below, we demonstrate the advantages of our formulation on two representative structure inference tasks: probabilistic matrix decomposition (\sect{sec:pmf}) and symbolic program parsing (\sect{sec:pp}). 

\section{Study I: Matrix Decomposition}
\label{sec:pmf}
\renewcommand{\myparagraph}[1]{\vspace{-10pt}\paragraph{#1}}

With the proliferation of structured probabilistic models such as binary latent factors~\citep{ghahramani2006infinite} and sparse coding~\citep{olshausen1996emergence,berkes2008sparsity}, people are getting more interested in the discovery of structures from data. In this section, we revisit the problem of structure inference for matrices~\citep{Grosse2012Exploiting}.

\subsection{Problem Formulation}

\begin{table}[t]
    \centering \small
\setlength{\tabcolsep}{1mm}{
    \begin{tabular}{lrcl}
        \toprule
        Low-rank & $G$ & $\rightarrow$ & $GG + G$\\ \midrule
        \multirow[t]{2}{*}{Clustering} & $G$ & $\rightarrow$ & $MG + G ~|~ GM^T $+$ G$\\
        & $M$ & $\rightarrow$ & $MG + G$\\ \midrule
        Markov chains & $G$ & $\rightarrow$ & $CG + G ~|~ GC^T + G$ \\ \midrule
        Gaussian scales & $G$ & $\rightarrow$ & $\exp(G) \circ G$ \\ \midrule
        \multirow[t]{2}{*}{Binary factors} & $G$ & $\rightarrow$ & $BG + G ~|~ GB^T + G$ \\
        & $B$ & $\rightarrow$ & $BG + G$ \\ \midrule
        Misc. & $M$ & $\rightarrow$ & $B$ \\
        \bottomrule
    \end{tabular}
}
    \caption{Production rules in the context-free DSL for structure inference. The left-most column indicates the type of the production rules.}
    \label{tab:matrix-production-rules}
    \vspace{-15pt}
\end{table}

We formally introduce the DSL for structural matrix decomposition via a motivating example. One of the simplest structures---Bayesian clustering---can be written as $F = MG + G$. Here, the symbol $M$ stands for a multinomial matrix, whose rows are sampled identically from a multinomial distribution. This matrix $M$ is multiplied by a Gaussian matrix $G$, whose rows are identically sampled from a Gaussian distribution. The conceptual meaning of the structure $MG + G$ could be interpreted as such: the row vector of the matrix stands for a stochastic choice of the cluster label, and the parameters of the multinomial distribution represent the probability of choosing each cluster. The first Gaussian matrix represents the center of each cluster. The last Gaussian matrix captures the \iid Gaussian noise. Other types of matrices in the DSL include $C$ (time-series Cumulative matrices) and $B$ (Binomial matrices).

\citet{Grosse2012Exploiting} proposed a context-free grammar to describe the structures of data matrices. \tbl{tab:matrix-production-rules} shows all production rules in the grammar. The inference always starts from a single symbol $G$, \ie, assuming all of the data items in the matrix to be \iid Gaussian.

\begin{table*}[t]
    \centering\small
    \setlength{\tabcolsep}{0pt}

    \begin{tabular}{lcccc}
    \toprule
        Data & Ground Truth & \citet{Grosse2012Exploiting}  & Mat2Seq  & \model (ours)\\
    \midrule
    Low-rank & $GG+G$ & Correct & Correct & Correct \\
    Clustering & $MG+G$ & Correct & Correct & Correct\\
    Binary latent features & $BG+G$& Correct & Correct & Correct\\
    Co-clustering & $M(GM^T+G)+G$& Correct & $MG+G$ & Correct \\
    Binary matrix factorization & $B(GB^T+G)+G$& Correct  & $BG+G$ & Correct\\
    BCTF & $(MG+G)(GM^T+G)+G$& Correct & $BG+G$ & Correct\\
    Sparse coding & $s(G)G+G$& Correct & $s(G)$ & Correct \\
    Dependent GSM & $s(GG+G)G+G$&  $s(G)G+G$ &  $s(G)$&  $s(G)G+G$ \\
    Random walk & $CG+G$& Correct & Correct & Correct\\
    Linear dynamical system & $(CG+G)G+G$& Correct  & $CG+G$ & $B(s(G)G+G)+G$\\ 
    \midrule
    Motion Capture - Level 1 & - & $CG+G$ & $CG+G$ & $CG+G$ \\
    Motion Capture - Level 2 & - & $C(GG+G)+G$ & $CG+G$ & $C(GG+G)+G${\big /}$CG+G$ \\
    Image Patch - Level 1 & - & $GG+G$ & $GG+G$ & $GG+G$ \\
    Image Patch - Level 2 & - & $s(G)G+G$ & $GG+G$ & $s(G)G+G$ \\
    Image Patch - Level 3 & - & {$\footnotesize{s(GG+G)G+G}${\big /}$s(G)G+G$} & ~~~~~$GG+G$~~~~~ & $s(G)G+G$ \\
    \bottomrule
    \end{tabular}
    \setlength{\tabcolsep}{6pt}
    \vspace{-3pt}
    \caption{Search results of our approach and several baselines. We denote $\exp(f(G))\circ G$ as $s(f(G))$. We run each model for three times with different random seeds. If the results produced by three runs disagree, we include all possible outcomes in the table, separated by $\big /$. See the main text for a detailed analysis.} 
    \label{tab:search-results}
    \vspace{-15pt}
\end{table*}

\subsection{Method}
The search-based algorithm \citet{Grosse2012Exploiting} requires evaluating all possible rules and ranking them with a proper metric to choose the one to be applied.
Intuitively, each production rule has a specific pattern of its matrices to be decomposed. For example, the data matrices having the ``clustering'' structure may have different patterns with matrices having the ``low-rank'' structure. Based on such observation, in \model, we adopt a Convolutional Neural Network (CNN) as a neural guider for the neurally-guided search. Requiring only limited synthetic data during training, the neural guider remarkably outperforms the algorithm-based exhaustive search of production rules in inference efficiency.

In detail, \model infers structures in a recursive manner following the paradigm shown as  Algorithm~\ref{alg:recursive-inference}. At each step, the neural guider and the symbolic decomposition algorithm work jointly: given the input data $D$, the neural guider predicts the production rule to be applied, and the symbolic decomposition algorithm decomposes the input data into multiple components. We use the decomposition algorithms and the metrics for ranking candidate structures in \citet{Grosse2012Exploiting}.

\myparagraph{Neural guider.} The neural guider is trained in the following way. Given a matrix $G$, the neural guider learns to distinguish a finer structure from all possible candidates, including $GG + G$, $MG + G$, $GM^T + G$, $CG + G$, $GC^T + G$, $BG + G$, $GB^T + G$ and $\exp(G) \circ G$.

\myparagraph{Training.} We generate synthetic data for training the neural guider on the fly.
For each data point, we first randomly sample a production rule (\eg, $G \rightarrow MG + G$) from the DSL. Then, we randomly generate a data matrix following the production rule (\eg, generate a multinomial matrix and two Gaussian matrices and compose them together) and use the underlying production rule as the label. The neural guider is trained to convergence on this dataset. It is then fixed for structure inference.

\myparagraph{Hyperparameters.}
We build our neural guider as an 8-layer convolutional neural network. 
The detailed architecture of the neural guider can be found in the supplementary material. To handle matrices of arbitrary sizes, we always pad the input matrices to $200\times200$ by adding zero entries. We also stack the padded input matrix, along the channel dimension, with a padding indicator matrix $P$: if the value at position $(i, j)$ belongs to the original input matrix, then $P[i, j] = 1$; otherwise, $P[i, j] = 0$.

We train the model with the Adam optimizer~\cite{kingma2014adam}. The hyperparameters for the optimizer are set to be $\beta_1=0.9, \beta_2=0.9, \alpha=10^{-4}$. The model is trained for $100{,}000$ iterations, with a batch size of $100$.

\begin{figure}[t]
    \centering
    \includegraphics[width=1.0\columnwidth,trim={0cm 0cm 0cm 0cm},clip]{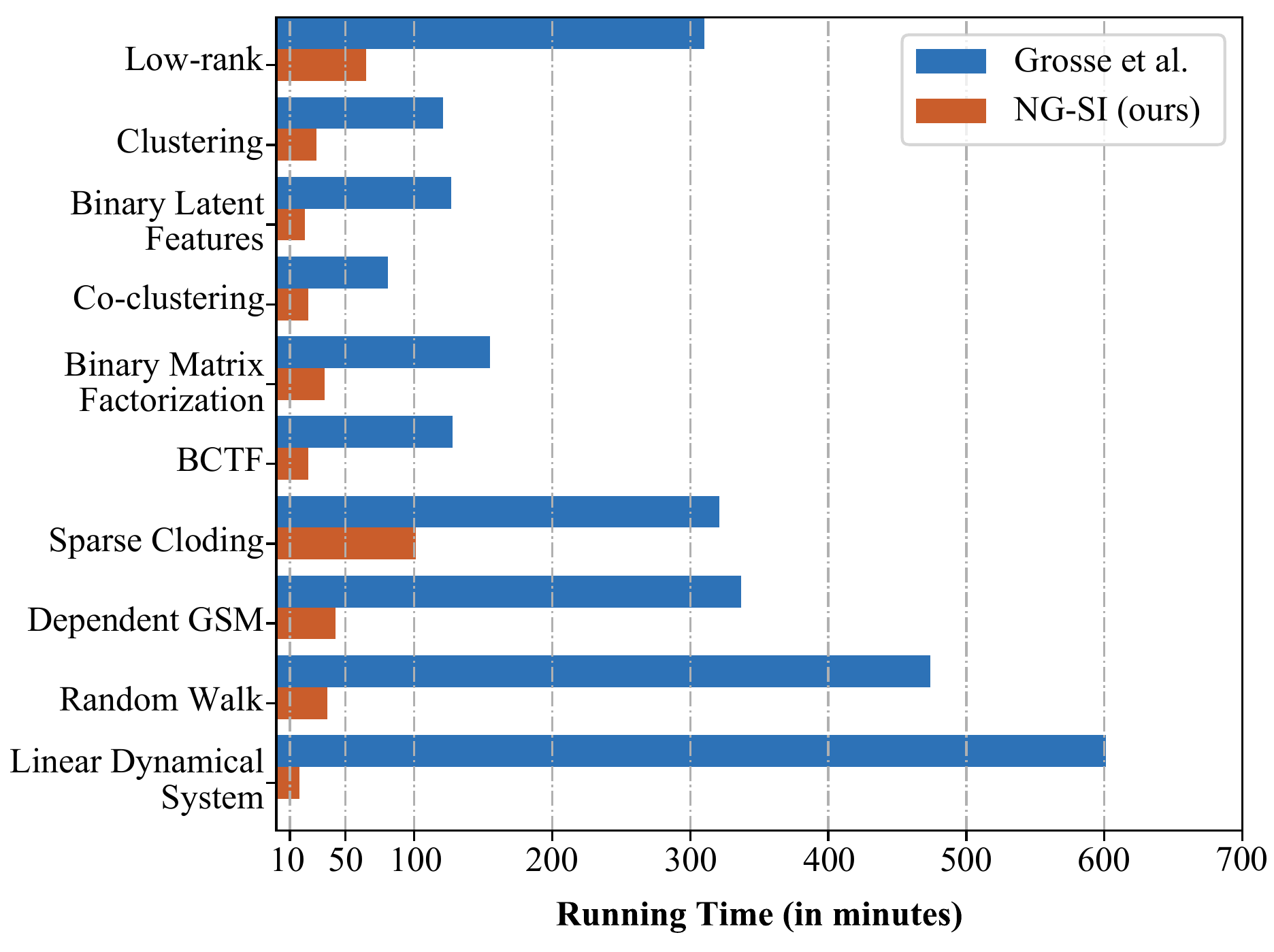}
    \vspace{-25pt}
    \caption{We compare the running time for structure inference of \cite{Grosse2012Exploiting} and \model. For all experiments, we set the standard deviation for the input noise to 1.  Remarkably, the speed of \model is $3\times \sim 39\times$ faster than the baseline.} 
    \label{fig:plot_time}
    \vspace{-15pt}
\end{figure}

\subsection{Experiments}
To evaluate the accuracy of our approach, we replicate the experiments in \citet{Grosse2012Exploiting}, including one synthetically generated dataset and two real-world datasets: motion capture and image patches. For the synthetic dataset, we generated matrices of size $200 \times 200$ from 10 models listed in \tbl{tab:search-results}. All models have a hidden dimension of 10, following \citet{Grosse2012Exploiting}.
The dataset of human motion capture \citep{hsu2005style, taylor2007modeling} consists of a person walking in a variety of styles. Each row of the data matrix describes the human pose in one frame, in the form of the person's orientation, displacement, and joint angles. The natural image patches dataset contains samples from the Sparsenet dataset proposed in \citet{olshausen1996emergence}. It contains 10 images of natural scenes (smoothed and whitened), from which 1,000 patches of size $12\times 12$ are selected and flattened as the rows of the matrix. We study the inferred structure with different search depth limits, varying from level 1 to level 3. Note that there is no groundtruth structure for such real-world datasets.

\myparagraph{Baselines.} 
Beside the search-based baseline in \citet{Grosse2012Exploiting}, we also implement a simple matrix-to-sequence model as a data-driven baseline. This model takes the data matrix as input and generates the structure of the matrix using a CNN-GRU model~\citep{vinyals2015show}. The baseline is trained on the same data as our neural guider.

\myparagraph{Accuracy.} Shown in \fig{fig:plot_time} and \tbl{tab:search-results}, our model successfully finds most of the optimal structures of synthetic data except for two of them: the dependent GSM and the linear dynamical system. Remarkably, the search process is accelerated with a multiplier of $3\times \sim 39 \times$. \model also generalizes well to real-world datasets. The inferred structures on both real-world datasets are consistent with the structures inferred by the search-based baseline.

For dependent GSM matrices, $(\exp(GG + G) \circ G)G + G$, the final structure determined by \model is the sparse coding model, $(\exp(G) \circ G)G + G$. This is a typical failure case as discussed in \citet{Grosse2012Exploiting}, since the variational lower-bound used for ranking candidate structures cannot distinguish two structures by a margin. We attribute the failure of the linear dynamical system case to the imbalance of training data: most production rules imply the independence of rows in the data matrix, while the rule $G \rightarrow CG + G$ does not. Thus, we see the misclassifications of structures that include this rule. In practice, the problem can be alleviated by techniques such as sampling more data with the $CG + G$ structure. 

For real-world datasets, the structures inferred by \model agree with the search-based baseline \citep{Grosse2012Exploiting} in most cases. Both methods show unstable results for complex structures; they sometimes fall back to a structure with a simpler but plausible form. We attribute this to the noises in real-world datasets, which affects the robustness of both the neural guider---for ranking production rules, and the variational lower-bound---for ranking inferred structures.

\myparagraph{Efficiency.} We empirically compare our algorithm against the original greedy search algorithm proposed by \citet{Grosse2012Exploiting}. We generate a $200 \times 200$ matrix from a dependent GSM model with $10$ latent dimensions. \tbl{tab:complexity-analysis} summarizes the running time needed for the search of structures of different depths. In general, our algorithm consistently speeds up the greedy search version by a factor of $8$.

\begin{table}[t]
    \centering\small
    \begin{tabular}{lccc}
    \toprule
        & Depth = 1 & Depth = 2 & Depth = 3 \\ 
    \midrule
        \citet{Grosse2012Exploiting} & 16min & 1h 32min & 5h 37min \\
        \model & 2min & 12min & 43min \\
        Speed up & $8\times$ & $7.67\times$ & $7.83\times$\\
    \bottomrule
    \end{tabular}
    \caption{Running time needed for the search of structures of different depths. Our algorithm consistently outperforms the baseline in efficiency. We ran all experiments on a machine with an Intel Xeon E5645 CPU and a GTX 1080 Ti GPU.} 
    \label{tab:complexity-analysis}
    \vspace{-10pt}
\end{table}

\myparagraph{Rule similarity discovery.} Our approach can be regarded as using an approximated probability distribution of structures conditioned on the input data to guide the structure search. Interestingly, we find that the learned distribution by the neural guider recovers the similarity between rules.

\begin{figure}[t]
  \centering 
  \includegraphics[width=\columnwidth]{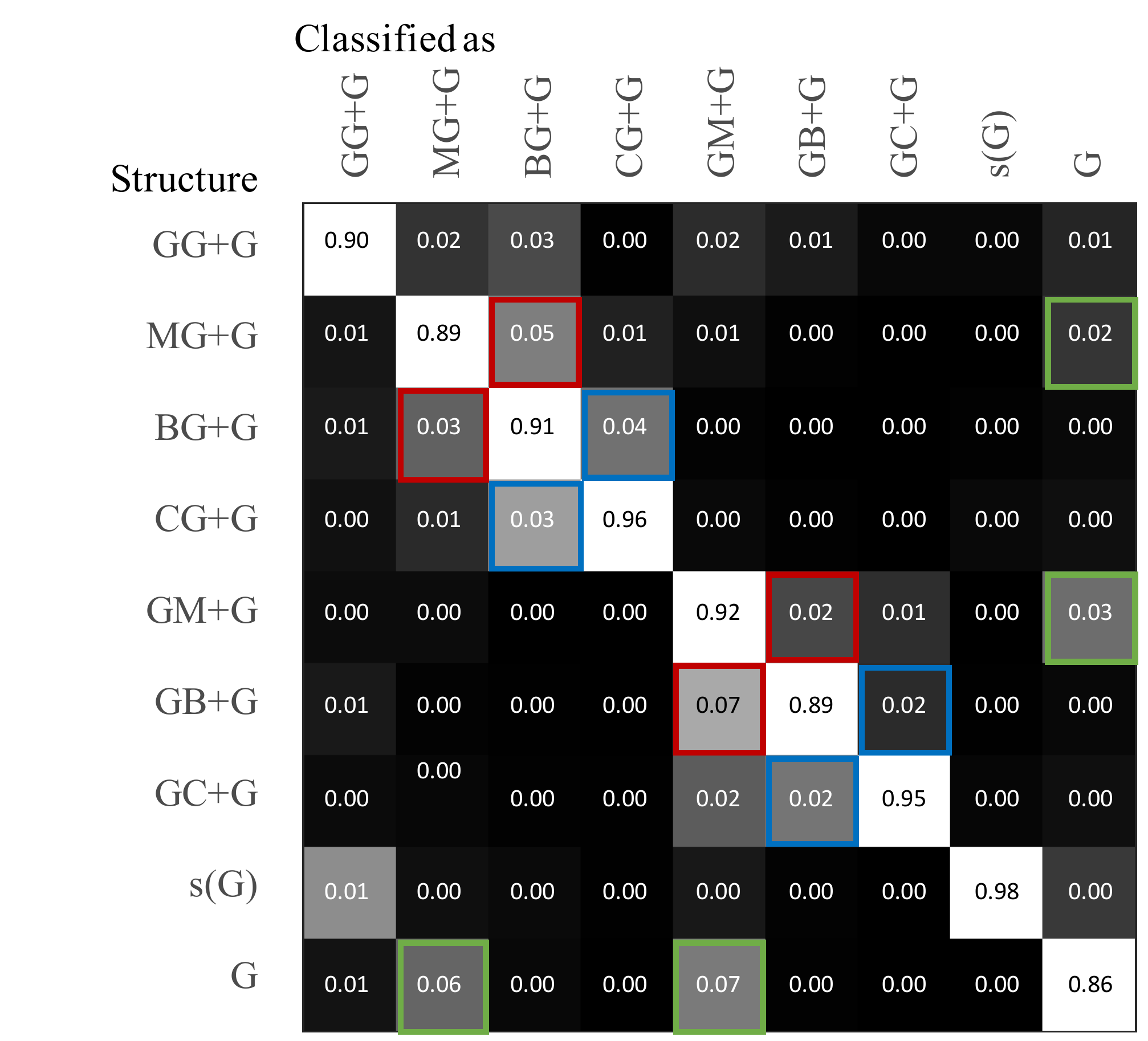}
  \vspace{-20pt}
  \caption{Visualization of the similarities between production rules for expanding a ``$G$'' node. The similarities are implicitly learned by our neural guider. We manually highlight some entries indication pairs of similar production rules. \textcolor{MyDarkRed}{$\Box$}: Clustering vs. Binary factor, \textcolor{MyDarkBlue}{$\Box$}: Binary factor vs. Markov Chain, and \textcolor{MyDarkGreen}{$\Box$}: Clustering (mixture of Gaussian) vs. Gaussian.}
  \label{fig:conf_mat}
  \vspace{-25pt}
\end{figure}

To visualize this, we first generate a dataset of matrices following different production rules, such as $G \rightarrow GG + G$ and $G \rightarrow MG + G$. All these rules start from a single matrix $G$. We then use the trained neural guider to predict the structure, \ie, the approximated probability distribution. We accumulate the output probabilities and visualize them as a matrix in \fig{fig:conf_mat}, where lighter entries indicate a higher probability of misclassification, or equivalently, a higher similarity between the two structures.

Ideally, the matrix should be symmetric, as it reflects the similarity between each two of the production rules. The empirical results support this intuition. Moreover, it recovers the similarity between some production rules. For example, $MG + G$ and $BG + G$ are similar (due to the similar binary structure of $M$ and $B$); $MG + G$ (mixture of Gaussian) and $G$ (pure Gaussian) are similar. These findings are consistent with human intuition.

\subsection{Application: Inspecting GANs}

The generative adversarial network (GANs) learns a transformation function $G_\theta$ (the generator) from a specific distribution (\eg, Gaussian) to the target data distribution \citep{Goodfellow2014Generative}. Such transformations are implemented as neural networks. Thus, it is usually difficult to interpret the generation process.

We view the generator of a GAN as a stack of distribution transformers, where each transformer is a single layer in the network. We show that it is possible to partially reveal the transformation process inside the GAN generator by detecting the structures of its intermediate features. By tracking these structures, we can obtain a better understanding of how GANs transform the distributions layer by layer.

As an example, we train an MLP-GAN to map a Gaussian distribution to randomly generated vectors from a set of dependent GSM distributions sampled from a common prior. The model is trained by the Wasserstein GAN-GP algorithm~\citep{Gulrajani2017Improved}. The generator is trained for $10{,}000$ iterations, and before each, the discriminator is trained for $4$ iterations. The architecture of the generator and the discovered structure of the intermediate features is summarized in \tbl{tab:gan-results}. In general, the trace of the structures is consistent with the natural compositional structure of the dependent GSM: $G \rightarrow GG+G \rightarrow s(G)G+G \rightarrow s(GG+G)G + G$.

\begin{table}[t]
    \centering\small
\setlength{\tabcolsep}{1mm}{
    \begin{tabular}{c}
        \toprule
        Gaussian noise $z$, dimension $=128$, structure: $G$\\
        \midrule
        Fully connected, dimension $=256$\\structure: $G$\\
        \midrule
        ReLU nonlinearity\\
        \midrule
        Fully connected, dimension $=256$
        \\structure: $MG+G$\\
        \midrule
        ReLU nonlinearity\\
        \midrule
        Fully connected, dimension $=256$\\structure: $GG+G$ {\big /} $BG+G$\\
        \midrule
        ReLU nonlinearity\\
        \midrule
        Fully connected, dimension $=256$\\structure: $s(G)G+G$\\
        \midrule
        ReLU nonlinearity\\
        \midrule
        Fully connected, dimension $=256$\\structure: $s(G)G+G$ or $s(GG+G)G+G$\\
        \bottomrule
    \end{tabular}
}
    \caption{The architecture and the discovered structures from the intermediate features of an MLP-GAN's generator. We run this experiment multiple times and show the top-ranked structures for each layer.}
    \label{tab:gan-results}
    \vspace{-5pt}
\end{table}

\renewcommand{\myparagraph}[1]{\vspace{-5pt}\paragraph{#1}}
\section{Study II: Program Parsing}
\label{sec:pp}
\renewcommand{\myparagraph}[1]{\vspace{-10pt}\paragraph{#1}}

The framework of neurally-guided structure inference can be naturally extended to other domains. In this section, we consider the task of program parsing from sequential data: given a discrete program written in a programming language, we want to translate it into a symbolic abstract syntax tree.

\subsection{Problem Formulation}

\begin{figure}[t!]
    \centering
    \includegraphics[width=\columnwidth]{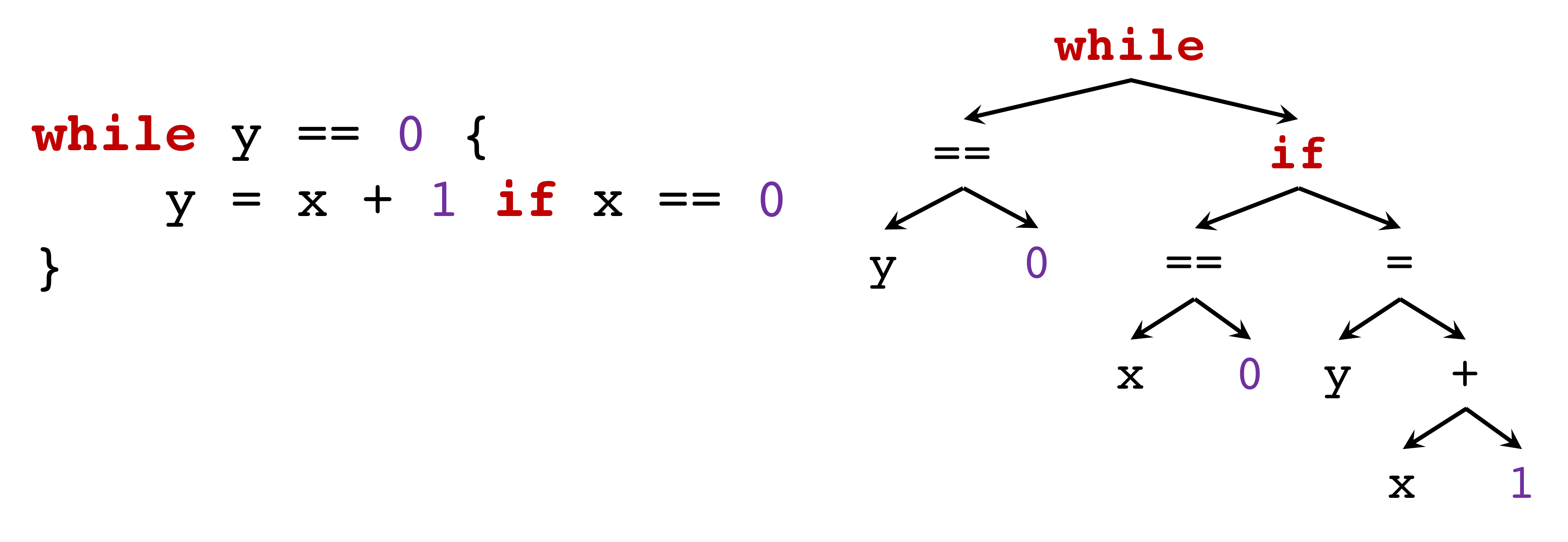}
    \vspace{-28pt}
    \caption{A sample code in the WHILE language (left) and its corresponding AST (right).}
    \label{fig:while-lang}
    \vspace{-18pt}
\end{figure}

For program parsing, we adopt the WHILE program language~\citep{chen2018towards} as our testbed. The WHILE language is defined by $73$ production rules. It covers most of the functionality of modern programming languages: arithmetic expressions, variable assignments, conditions, and loops. Our goal is to translate a program written in the WHILE language into a hierarchical abstract syntax tree (AST). \fig{fig:while-lang} shows a sample code written in the WHILE language and its corresponding AST. Ideally, after learning from a limited number of programs and the AST labels, the algorithms should generalize to parse longer programs or more complex programs, \ie, programs with a deeper AST. As we show later, such generalizability is difficult for many data-driven algorithms such as sequence-to-sequence models~\citep{Sutskever2014Sequence}.

\subsection{Model}

\begin{figure*}[t!]
  \centering
  \subfigure[Depth = 6]{\includegraphics[width=0.32\linewidth]{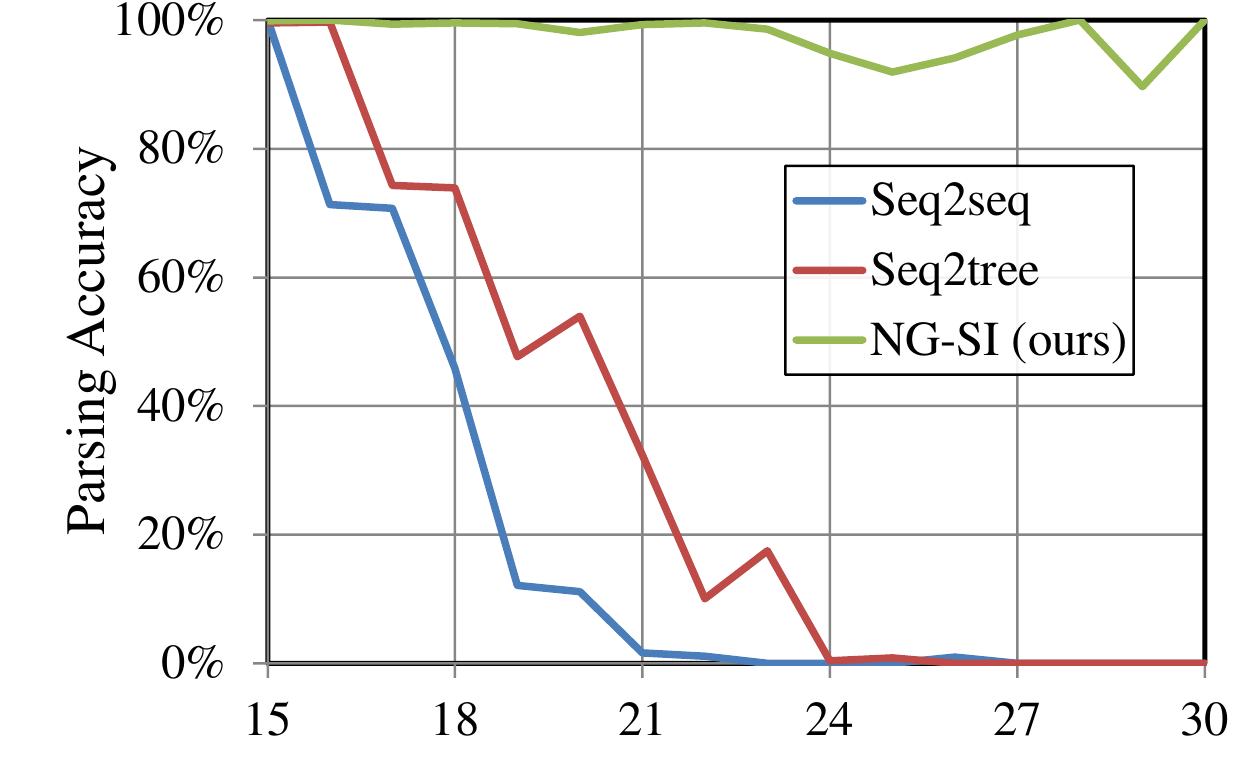}}~~~~
  \subfigure[Depth = 7]{\includegraphics[width=0.32\linewidth]{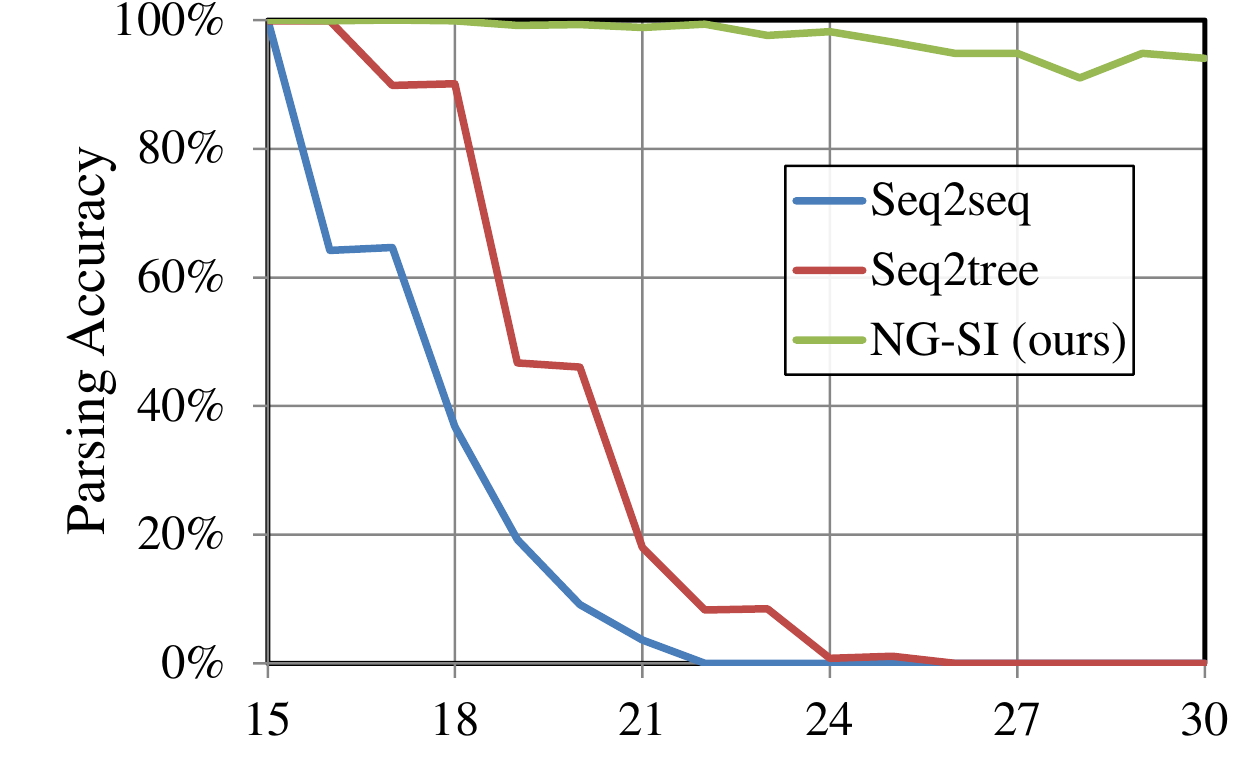}}~~~~
  \subfigure[Depth = 8]{\includegraphics[width=0.32\linewidth]{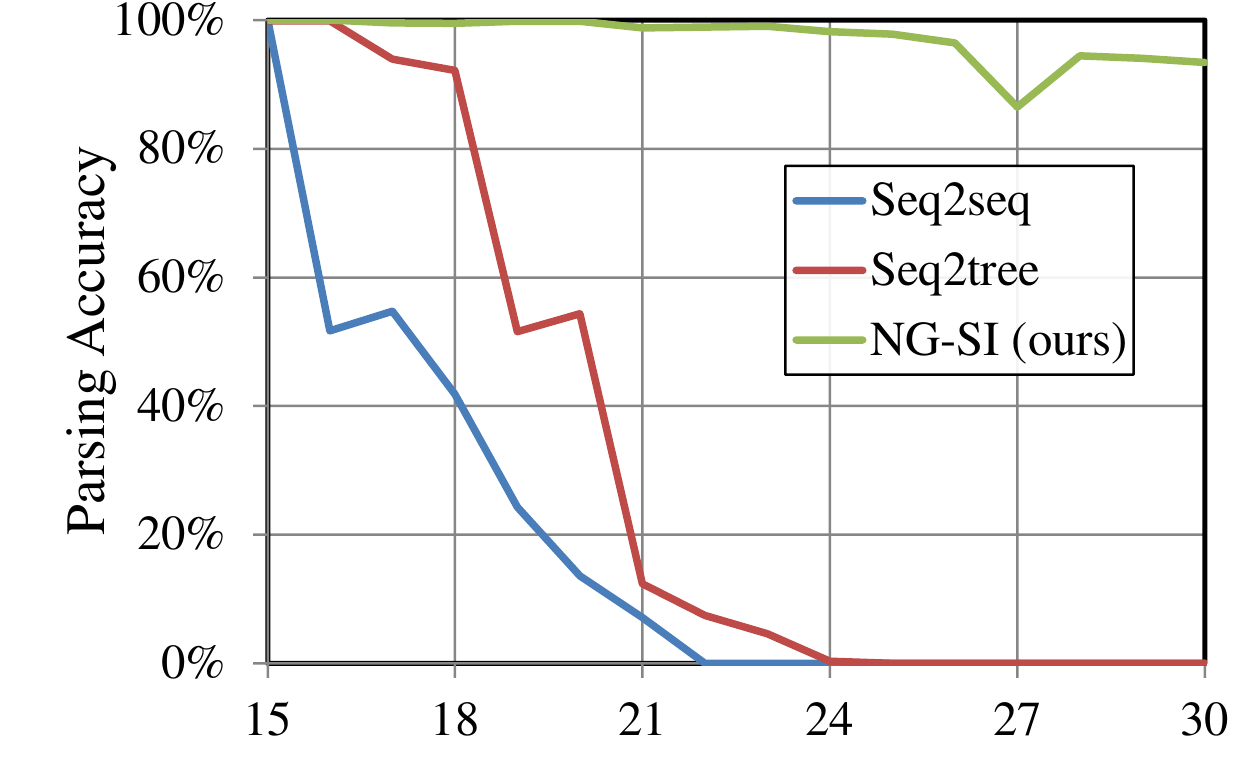}}
  
  \subfigure[Depth = 9]{\includegraphics[width=0.32\linewidth]{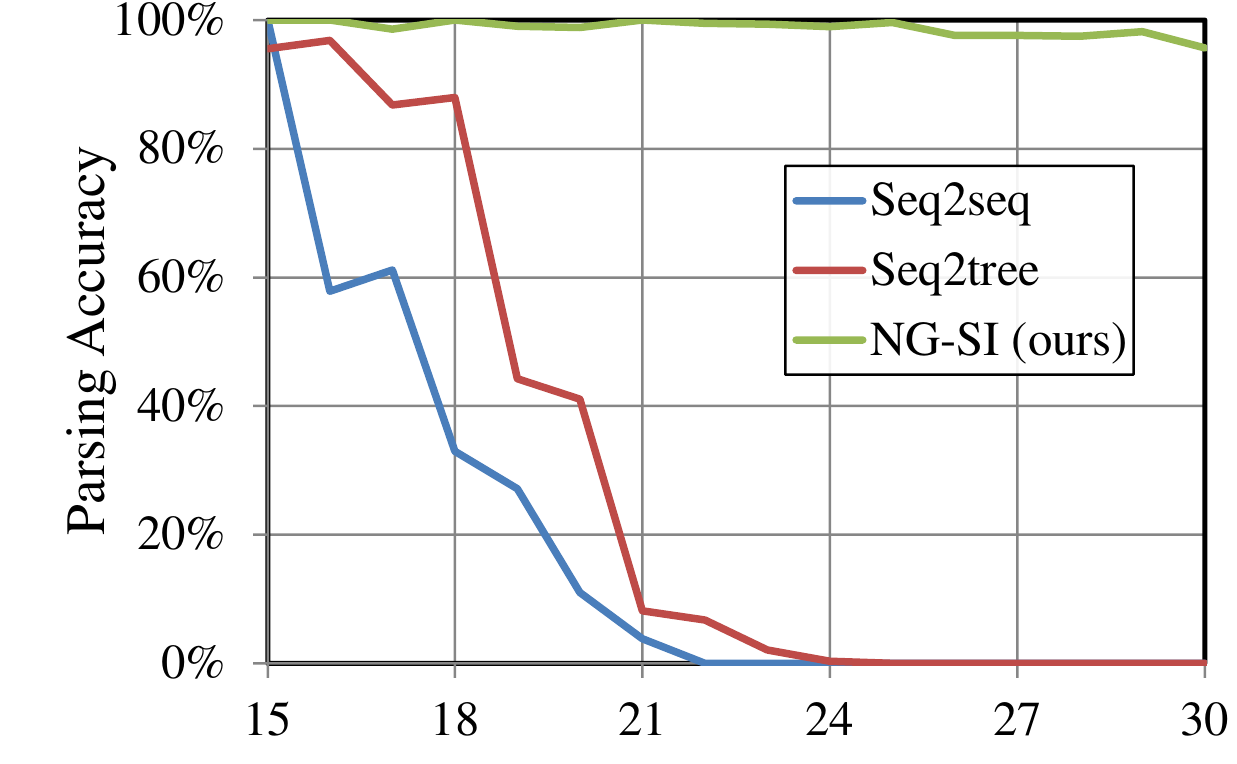}}~~~~
  \subfigure[Depth = 10 \label{fig:parsing_acc:10}]{\includegraphics[width=0.32\linewidth]{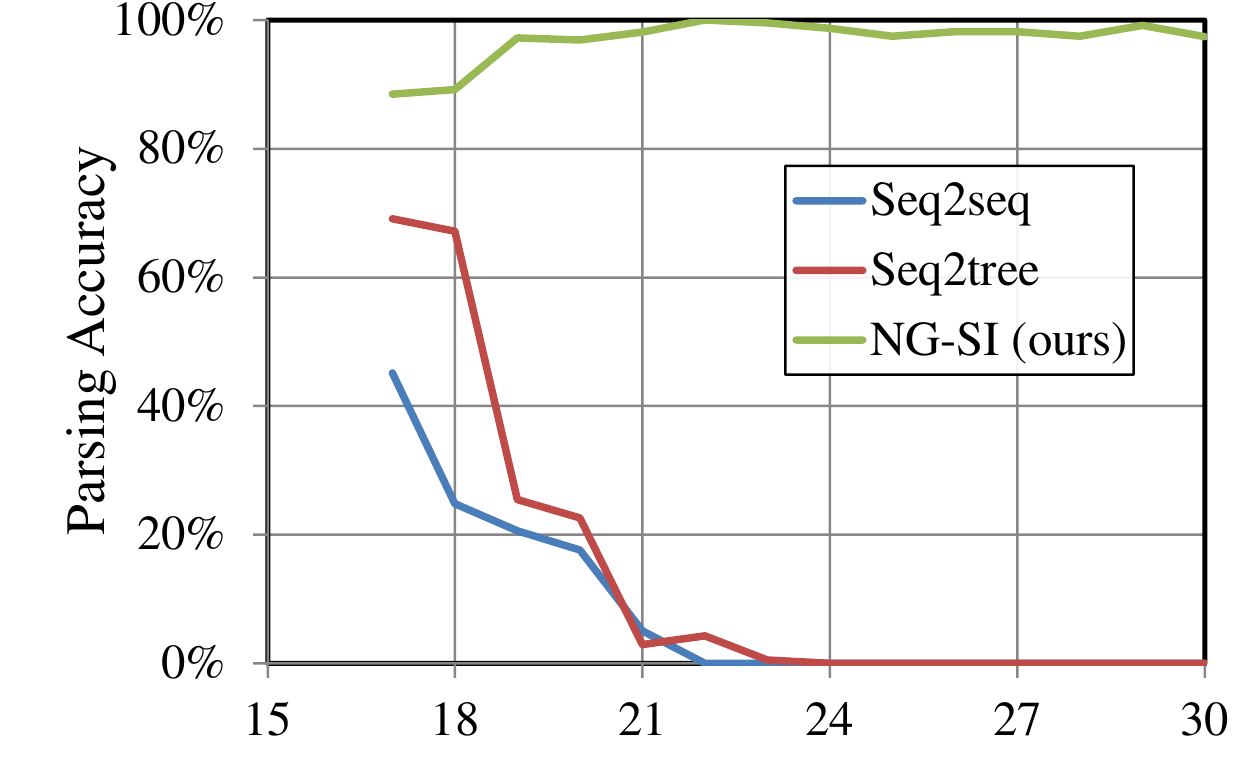}}~~~~
  \subfigure[Depth = 11 \label{fig:parsing_acc:11}]{\includegraphics[width=0.32\linewidth]{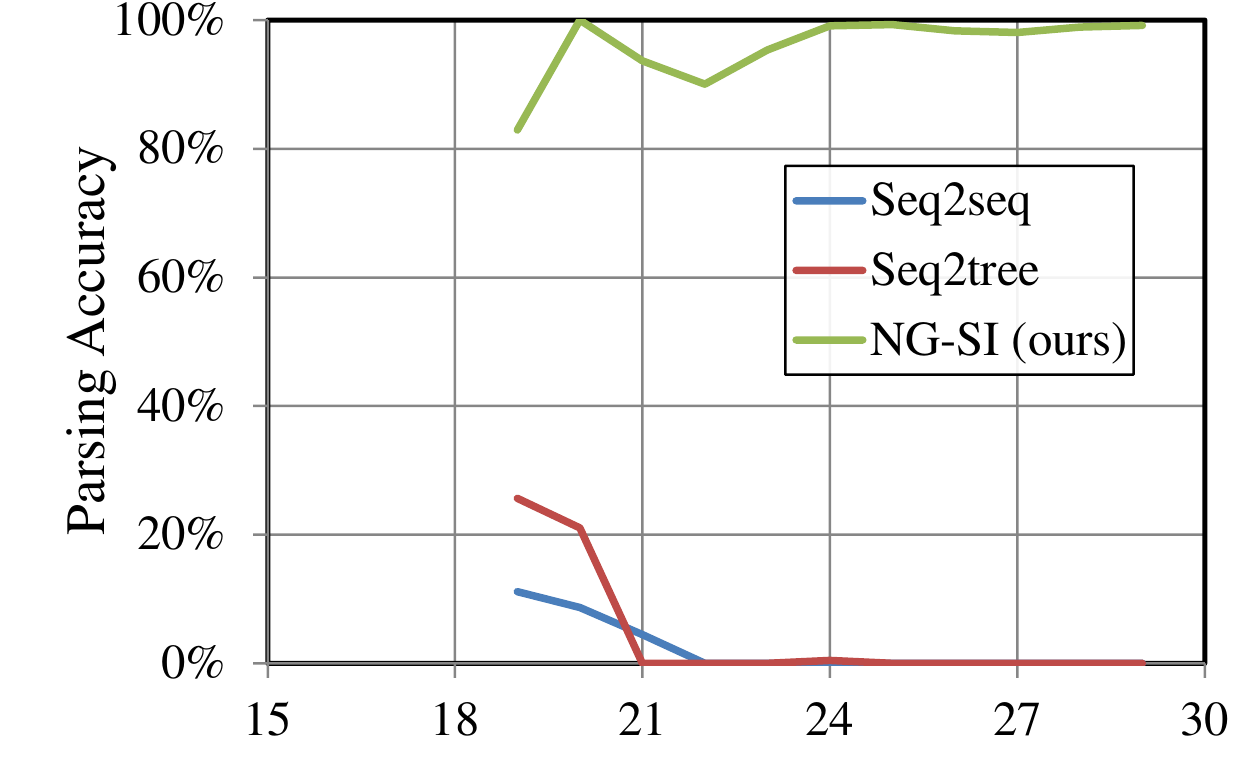}}
  \vspace{-15pt}
  \caption{The performance of \model and the baselines (Seq2Seq and Seq2Tree) on test programs with different AST depths ($6$ to $11$) and different lengths ($15$ to $30$). Deep programs (depth $\geq 10$) have a minimal length larger than 15: Figure~\ref{fig:parsing_acc:10} and Figure~\ref{fig:parsing_acc:11}}
  \label{fig:parsing_acc}
  \vspace{-10pt}
\end{figure*}

Following the same hybrid search paradigm as presented in Algorithm~\ref{alg:recursive-inference}, we build a neurally-guided program parser. We keep the design of recursive inference: at each step, the neural guider infers the production rule based on the input code. Then we decompose the code into components (such as the conditions and the body for an if-statement). Thus, the data are represented as code strings. Here, we use hand-coded algorithms for code decomposition.

Because the neural guider determines only the top-level production rule for the input code, intuitively, it does not require specific architecture designs. In our experiments, we implement it as a combination of a GRU encoder and a linear classifier. The input code string is first fed into the GRU encoder; then the classifier takes the last hidden state of the encoded string as its input.

\myparagraph{Training.} We randomly generate training samples based on the context-free grammar of the WHILE language. Roughly, starting from the start symbol of the grammar, we randomly apply a number of production rules on non-terminal symbols. For each generated program, all of its sub-strings corresponding to one of the sub-trees in the AST are used as the training data. To test the generalizability of the models, we restrict the depth of the AST of the training examples and the length of the programs to be less than $9$ and $15$, respectively. The learned model is tested on longer programs or more complex programs (\ie, with a deeper AST).

\myparagraph{Hyperparameters.} We adopt a unidirectional GRU with a hidden dimension of 256 as the code string encoder for production rule selection. We train the model using the Adam optimizer, with hyperparameters $\beta_1=0.9, \beta_2=0.9, \alpha=10^{-4}$. The batch size is set to $64$. We perform curriculum learning similar to the training process described in \citet{chen2018towards}, where the model is trained with programs of gradually increasing length and depth. The shortest program has a length of $5$, while the longest has $15$. We repeat the curriculum learning process three times for training the neural guider.

\subsection{Experiments}

\paragraph{Baselines.}
We implement a sequence to sequence (Seq2Seq) model~\citep{Sutskever2014Sequence} with attention and a Seq2Tree model~\citep{dong2016language} model with attention as the data-driven baselines for AST inference.

The Seq2Seq model with attention uses GRUs as the encoder and the decoder, both with a hidden dimension of $256$. We slightly modify the output format of the Seq2Seq baseline to support the tree-structured output. Specifically, we perform a pre-order traversal of the AST and use the traversal order of all nodes as the label for training \citep{vinyals2015grammar}.  

The Seq2Tree model uses GRU as the encoder, with a hidden dimension of $256$ and a hierarchical tree decoder. It generates the AST in a top-down manner. Starting from the starting symbol, the decoder iteratively expands a non-terminal symbol with a production rule predicted by the decoder network.

For training both baseline models, we use the teacher forcing method and the Adam optimizer with the same hyperparameters as the neural guider.

\myparagraph{Results.}
\fig{fig:parsing_acc} shows the results. We evaluate the performance of \model and the baseline models on programs with different AST depths ($6$ to $11$) and different lengths ($15$ to $30$). \fig{fig:parsing_acc} shows that \model robustly infers the hierarchical AST from more complex and longer programs than training examples. By contrast, the performance of the purely data-driven baseline drops significantly as the complexity or the length of the program grows beyond training examples.
It is worth noting that \model is remarkably robust \wrt the program depth. Although the model has never seen programs with depth 11 during training, it achieves an accuracy $\ge$ 90\% during inference. In contrast, the accuracies of all baselines are $\le$ 20\%.

As for the running time, Seq2Seq, Seq2Tree, and our proposed NG-SI give prediction in less than 1s per instance.
We also compare our model with an exhaustive search baseline. Specifically, it uses iterative deepening depth-first search for programs. The search stops when it finds an AST that exactly reconstructs the input program. The algorithm runs in a single thread on a machine with an Intel Core i7-8700 4.0GHz and 16G RAM, and the running time limit is 1 hour. With a fixed program depth = 6, the search-based baseline achieves perfect accuracy on programs with length $\le$ 24 but exceeds the running time limit when length $> 24$.

\renewcommand{\myparagraph}[1]{\vspace{-5pt}\paragraph{#1}}
\section{Conclusion}
We have proposed the \modelfull (\model), a hybrid algorithm for structure inference that keeps the advantages of both search-based and data-driven approaches. The key idea is to use a neural network to learn to guide a hierarchical, layer-wise search process. The data-driven module enables efficient inference: it recursively selects the production rules to build the structure, so that only a small number of nodes in the search tree need to be evaluated. The search-based framework of \model ensures robust inference on test data with arbitrary complexity. Results on probabilistic matrix decomposition benchmarks and program parsing datasets support our arguments.

\myparagraph{Acknowledgements.}
We thank Xinyun Chen for helpful discussions and suggestions.
This work was supported in part by the Center for Brains, Minds and Machines (CBMM, NSF STC award CCF-1231216), ONR MURI N00014-16-1-2007, and Facebook.

\bibliography{reference,struct}
\bibliographystyle{icml2019}

\newpage
\quad
\newpage

\includepdf{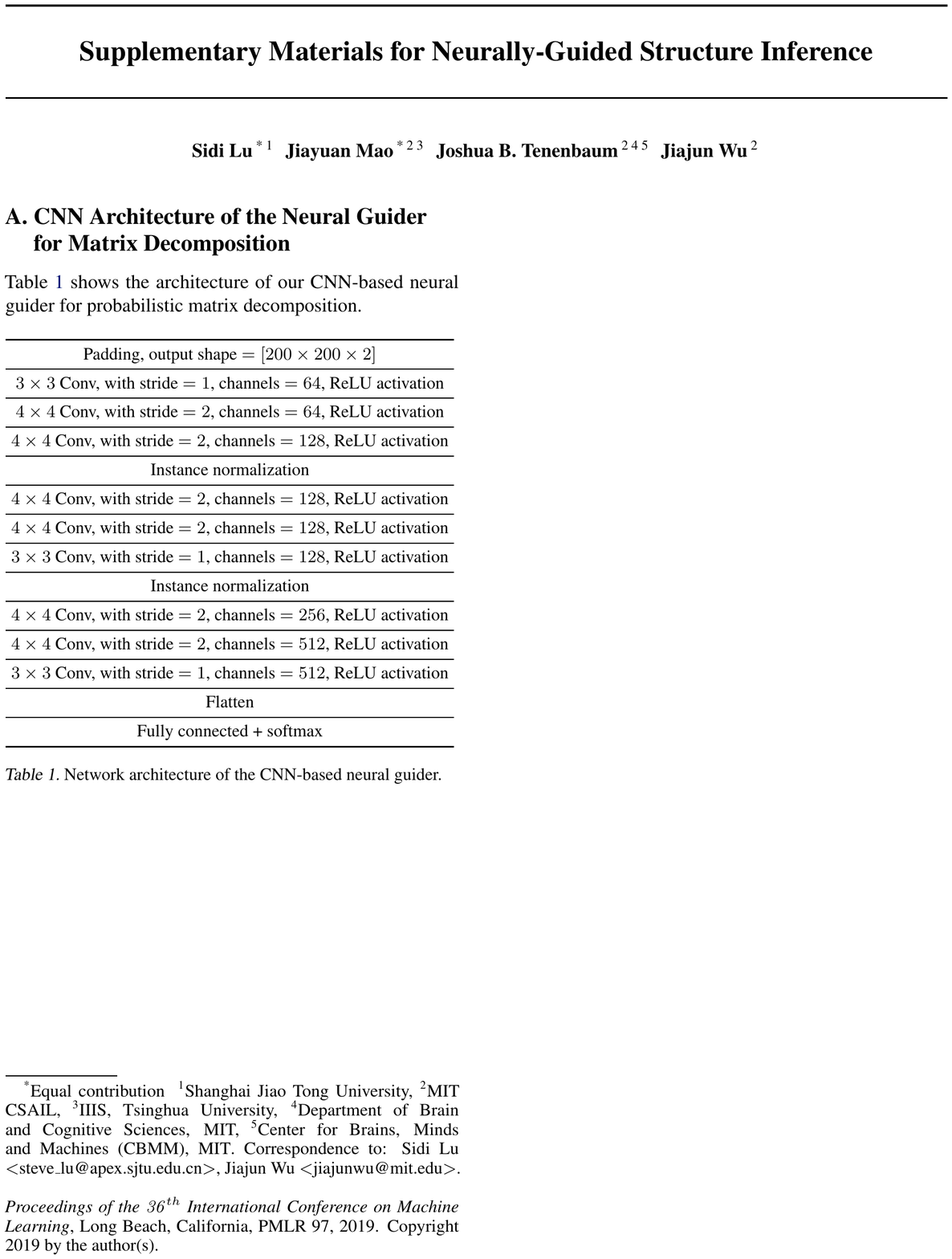}

\end{document}